\pdfoutput=1

\documentclass[11pt]{article}

\usepackage[]{EMNLP2022}

\usepackage{times}
\usepackage{latexsym}

\usepackage[T1]{fontenc}

\usepackage[utf8]{inputenc}

\usepackage{microtype}

\usepackage{url}
\usepackage{graphicx} 
\usepackage{float} 
\usepackage{booktabs} 
\usepackage{amsmath}
\usepackage{amsfonts} 

\usepackage{inconsolata}

\title{Let Me Check the Examples: Enhancing Demonstration Learning \\ via Explicit Imitation}

\author{Sirui Wang, Kaiwen Wei, Hongzhi Zhang, Yuntao Li and Wei Wu \\
    Meituan Inc., Beijing, China \\
  \texttt{\{wangsirui,weikaiwen,zhanghongzhi03,liyuntao04,wuwei30\}@meituan.com} \\
}

\begin{document}
\maketitle
\begin{abstract}
    Demonstration learning aims to guide the prompt prediction via providing answered demonstrations in the few shot settings. 
    Despite achieving promising results, existing work only concatenates the answered examples as demonstrations to the prompt template (including the raw context) without any additional operation, neglecting the prompt-demonstration dependencies.
    Besides, prior research found that randomly replacing the labels of demonstrations marginally hurts performance, illustrating that the model could not properly learn the knowledge brought by the demonstrations. 
   Inspired by the human learning process, in this paper, we introduce \textbf{Imitation DEMO}nstration Learning (Imitation-Demo)  
   to strengthen demonstration learning via explicitly imitating human review behaviour, which includes: 
   (1) contrastive learning mechanism to concentrate on the similar demonstrations.
   (2) demonstration-label re-prediction method to consolidate known knowledge.
    Experiment results show that our proposed method achieves state-of-the-art performance on 11 out of 14 classification corpora. Further studies also prove that Imitation-Demo strengthen the association between prompt and demonstrations, which could provide the basis for exploring how demonstration learning works.
\end{abstract}

\section{Introduction}
\begin{figure*}[!t] 
	\centering
	\includegraphics[width=15.5cm]{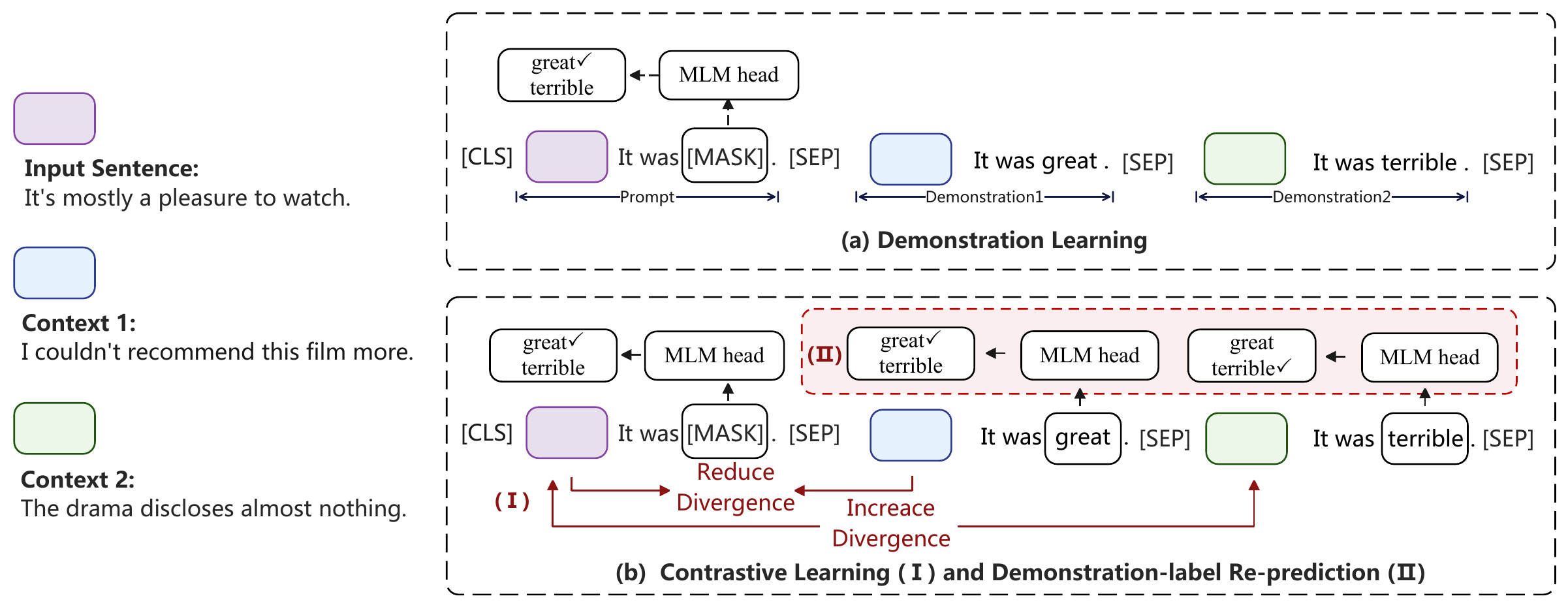}
	\caption{The overview of our proposed Imitation-Demo:
	(a) Conventional demonstration learning simply concatenate the demonstrations to the prompt. (b) Imitation-Demo reinforces the dependencies between prompt and demonstrations via contrastive learning (\uppercase\expandafter{\romannumeral1}) and demonstration-label re-prediction (\uppercase\expandafter{\romannumeral2}). 
	For brevity, all sentences and contexts from demonstrations are represented by coloured boxes (illustrated in the left part). Best view in colours. }
	\label{model} 
\end{figure*}

Prompt-based learning typically works by modifying the input into cloze-style prompt templates and using the masked language models (MLMs) to complete the unfilled information in probabilistic.
It has achieved promising performance in various NLP tasks \cite{DBLP:conf/eacl/SchickS21,DBLP:conf/emnlp/LesterAC21,DBLP:journals/corr/abs-2108-02035}, especially in low-resource settings \citep{DBLP:conf/naacl/ScaoR21}. 
A promising prompt engineering category is \textit{demonstration learning} \cite{DBLP:conf/acl/GaoFC20,liu2021pre}, which seeks to provide a few answered samples as demonstrations to assist prompt prediction. As shown in Fig. \ref{model} (a), the demonstration learning method concatenates the answered demonstrations per category to the prompt, and seeks to classify the $[MASK]$ token as $great$, indicating a $positive$ prediction result based on a label-to-word mapping.


The intuition of demonstration learning is that samples with similar expressions or content can provide repetitive patterns \cite{liu2021pre}.
However, \citet{DBLP:journals/corr/abs-2202-12837} point out that replacing gold demonstration labels with random labels marginally hurts performance. This finding is counter-intuitive and illustrates that the model could not comprehensively refer to the knowledge brought by the demonstrations in an implicit way. 
We attribute this problem to the fact that existing methods simply concatenate the answered demonstrations to the prompt template without any additional operation, ignoring the dependencies between prompt and demonstrations. 

To overcome this limitation, we rethink how human beings learn from demonstrations. 
When faced with a new challenging question, they typically
 (1) look for the most similar example to the question first, and then (2) reply to the question according to the answering steps of the retrieved example. 
Humans tend to strengthen the learning process through review strategies, i.e., finding a better solution to select similar examples and re-answering the questions of examples to consolidate known knowledge. 
Inspired by this, likewise, the interactions between prompt and demonstrations could also be reinforced by imitating the human reviewing process for demonstration learning. 


In this paper, we propose a simple-yet-effective version of demonstration learning, named \textbf{Imitation DEMO}nstration Learning (Imitation-Demo) to explicitly strengthen the two sub-steps of demonstration learning via human-like review. 
Specifically, for accurately locating similar samples, we introduce a contrastive learning mechanism \cite{DBLP:conf/icml/ChenK0H20, DBLP:conf/iclr/RobinsonCSJ21} to reorganize demonstrations by
reducing the divergences of demonstration contexts among the same category while increasing those divergences between different categories.
In addition, for solidifying known knowledge, we leverage a demonstration-label re-prediction method to emphasize the positions of the answers in demonstrations.
Even without introducing new parameters or any prediction computation, our proposed method achieves state-of-the-art performance on 11 out of 14 classification corpora. Compared to the strong baseline model LM-BFF \cite{DBLP:conf/acl/GaoFC20}, Imitation-Demo achieves 1.11 points averaged improvement on the 14 datasets. Further study also shows that Imitation-Demo strengthens the association between prompt and demonstrations, which could provide the basis for exploring how demonstration learning works.

\section{Methodology}  
	

\textbf{Demonstration Learning. }
As illustrated in Fig. \ref{model} (a), firstly, we leverage the pre-trained SBERT \cite{DBLP:conf/emnlp/ReimersG19} to retrieve the demonstration (including context $x^{(k)}$ and label $y^{(k)}$) for the $k$-th category that has maximum semantic similarity to the raw prompt context. 
The prompt template $x^{prompt}$ consists of input sentence $x^{in}$ and template $x^{temp}$ containing mask token,  i.e., $x^{prompt}=[x^{in},x^{temp}]$. Then, the demonstrations are concatenated to the input prompt. 
After that, we convert the concatenated input sentence $x^{in}$ to hidden vectors $\mathbf{h}^{in}$. The goal of demonstration learning is to predict $y^{mask}$ at the $[MASK]$ position from the hidden state of mask $\mathbf{h}^{mask}$ via MLM head. The overall process is optimized by cross-entropy loss and could be formulated as\footnote{Please refer to \citet{DBLP:conf/acl/GaoFC20} for more details of demonstration learning. }:
\begin{align}\begin{aligned}
\!\!x^{in}=[x^{prompt}, (x^{(1)},y^{(1)}),...,&(x^{(K)},y^{(K)})] \\
\!\!\mathbf{h}^{in} = \text{SBERT}(x^{in})& \\
\!\!p\left(y^{mask} \mid  x_{\mathrm{in}}\right)  = \text{MLM}(&\mathbf{h}^{mask}) \\
\!\!\mathcal{L}_{mask} = \text{CE}  (p\left(y^{mask} \mid  x_{\mathrm{in}} \right),& \hat{Y}^{(mask)} )
\end{aligned}\end{align}
where [.. ,.. ,..] denotes concatenating diverse parts with sentence separator $[SEP]$. $K$ is the number of categories. CE is short for cross-entropy loss, and $\hat{Y}^{(mask)}$ is the ground-truth labels from the pre-defined label-to-word mapping.
\\\textbf{Demonstration Reorganization via Contrastive Learning. }
In demonstration learning, it is crucial to decide from which known demonstrations to select the repetitive patterns.
Therefore, we introduce a contrastive learning mechanism to imitate human review behaviour by reorganizing the demonstrations based on their contexts. 
As shown in Fig. \ref{model} (b)(\uppercase\expandafter{\romannumeral1}), we treat the demonstration contexts with identical categories to the input prompt as positive samples, and the others are regarded as negative ones. By pulling in positive samples and pulling out negative samples, the model could select the most relevant sample among the given demonstrations more precisely. In the experiment, we apply mean-pooling operations on the hidden states of positive, negative demonstration contexts $\mathbf{h}^{+}$, $\mathbf{h}^{-}$, and input sentence $\mathbf{h}^{in}$, obtaining the sentence representations $\mathbf{s}^{+}$, $\mathbf{s}^{-}$, and $\mathbf{s}^{in}$. 
Inspired by \citet{DBLP:conf/iclr/RobinsonCSJ21} in computer vision, we introduce HCL loss to 
ensure intra-class compactness while increasing inter-class distances:
\begin{equation}
    \!\!\!\mathcal{L}_{context} \!=\! \mathbb{E}\left[-\log \frac{e^{{s}^{in} \cdot s^{+}}}{e^{{s}^{in}\cdot  s^{+}}+ \sum_{i=1}^{N} e^{{s}^{in} \cdot  s^{-}}}\right]
\end{equation}
where $\cdot$ is the dot product operation, $N$ is the number of negative contexts in the task, and $\mathbb{E}\left[..\right]$ denotes calculating the mean value.\\
\textbf{Demonstration-label Re-prediction. } We further utilize a demonstration-label re-prediction method to mimic human review behaviour by 
recovering the labels from all the given demonstration contexts. 
Specifically, the target of our model is not only to identify the category of $[MASK]$ token, but also to classify the tokens located in demonstration label positions. 
As illustrated in Fig.\ref{model} (b)(\uppercase\expandafter{\romannumeral2}), the model is required to predict $y^{great}$ or $y^{terri}$ (i.e., $great$ or $terrible$) based on the hidden states $\mathbf{h}^{great}$ or $\mathbf{h}^{terri}$ at corresponding label positions: 
\begin{equation}
\begin{split}
    p\left(y^{great} \mid x_{\mathrm{in}}\right) = \text{MLM}(\mathbf{h}^{great}) \\
    p\left(y^{terri} \mid x_{\mathrm{in}}\right) = \text{MLM}(\mathbf{h}^{terri}) \\
\end{split}
\end{equation}

The cross-entropy loss $\mathcal{L}_{great}$ and $\mathcal{L}_{terri}$ are calculated for each demonstration label, then we get the sum loss $\mathcal{L}_{label}$:
\begin{equation}
\begin{split}
    \mathcal{L}_{great} = \text{CE} & (p\left(y^{great} \mid x_{\mathrm{in}}\right), \hat{Y}^{(demo)}) \\
    \mathcal{L}_{terri} = \text{CE} & (p\left(y^{terri} \mid x_{\mathrm{in}}\right), \hat{Y}^{(demo)}) \\
    \mathcal{L}_{label} &= \mathcal{L}_{great} + \mathcal{L}_{terri}
\end{split}
\end{equation}
where $\hat{Y}^{(demo)}$ is the ground-truth labels at diverse demonstration label positions. 

\begin{table}[b!]
\centering
\scalebox{0.8}{
\begin{tabular}{lcccc}
\toprule[1pt]
        & MRPC       & SNLI       & SST-2      \\ \hline
Imitation-Demo     & 80.8 (3.2) & 80.0 (3.3) & 93.1 (0.5) \\
LM-BFF* & 79.7 (3.2) & 77.8 (0.6) & 92.1 (1.5) \\
Imitation-Demo*     & 74.4 (9.2) & 76.0 (5.2) & 91.0 (1.3) \\  \bottomrule[1pt]
\end{tabular}
}
\caption{Results when using demonstrations with random labels. * denotes trained with random labels. }
\label{test:random}
\end{table}

Similar operations can be performed for classification tasks with multiple categories. The overall loss of Imitation-Demo is defined as follows:
\begin{equation}
    \mathcal{L}  = \mathcal{L} _{mask} + \alpha \mathcal{L} _{label}+ \beta \mathcal{L} _{context}
\end{equation}
where $\alpha$, $\beta$ are weight coefficients for different components.



\begin{table*}[t!]
\scalebox{0.79}{
\centering
\begin{tabular}{lccccccc}
\toprule[1pt]
                               & \begin{tabular}[c]{@{}c@{}}SST-2\\ $($acc$)$ \end{tabular}     & \begin{tabular}[c]{@{}c@{}}SST-5\\ $($acc$)$\end{tabular}     & \begin{tabular}[c]{@{}c@{}}MR\\ $($acc$)$ \end{tabular}                            & \begin{tabular}[c]{@{}c@{}}CR\\ $($acc$)$ \end{tabular}        & \begin{tabular}[c]{@{}c@{}}MPQA\\ $($acc$)$ \end{tabular}      & \begin{tabular}[c]{@{}c@{}}Subj\\ $($acc$)$ \end{tabular}      & \begin{tabular}[c]{@{}c@{}}TREC\\ $($acc$)$ \end{tabular}                           \\ \hline
Majority                       & 50.9      & 23.1      & 50.0                          & 50.0      & 50.0      & 50.0      & 18.8                          \\
Prompt-based zero-shot         & 83.6      & 35.0      & 80.8                          & 79.5      & 67.6      & 51.4      & 32.0                          \\
“GPT-3” in-context learning    & 84.8 (1.3) & 30.6 (0.9) & 80.5 (1.7)                     & 87.4 (0.8) & 63.8 (2.1) & 53.6 (1.0) & 26.2 (2.4)                     \\ 
Fine-tuning    & 81.4 (3.8) & 43.9 (2.0) & 76.9 (5.9)                     & 75.8 (3.2) & 72.0 (3.8) & 90.8 (1.8) & 88.8 (2.1)                     \\
P-tuning    & 92.2 (0.4) & - & 86.7 (1.2)                     & 91.8 (1.1) & - & 90.3 (2.2) & 86.3 (4.5)                     \\
DART  & \textbf{93.5} (0.5) & - & 88.2 (1.0)                     & 91.8 (0.5) & -  & 90.7 (1.4) & 87.1 (3.8)                     \\ 
Li's & 92.8 (0.6) & 50.7 (2.9)                     & \textbf{89.4} (0.8) & 90.5 (2.2)  & 83.2 (1.4) & 92.1 (0.7) & 87.2 (3.8) \\
EFL $^{\heartsuit}$ & 91.1 (1.5) & 41.8 (1.6)                     & 85.7 (3.7) & 87.7 (5.4)  & 75.8 (4.8) & 91.7 (1.8) & 88.1 (2.3) \\
LM-BFF $^{\heartsuit}$ & 92.2 (1.4) & 51.2 (1.6) & 88.2 (0.9)                     & 91.8 (1.5) & 85.5 (4.2) & 90.9 (1.9) & \multicolumn{1}{l}{87.6 (4.8)} \\
Imitation-Demo (ours) & 93.1 (0.5) & \textbf{52.3} (0.6) & 89.1 (1.0)                     & \textbf{91.8} (0.7) & \textbf{87.7} (1.2) & \textbf{92.4} (1.1) & \textbf{89.1} (3.2)      \\
 \hline
Prompt-based Fine-tuning (man) $^{\heartsuit}$         & 92.6 (0.5) & 47.4 (2.5) & 87.0 (1.2)                     & 90.3 (1.0) & 84.7 (2.2) & 91.2 (1.1) & \multicolumn{1}{l}{84.8 (5.1)} \\
\quad + demonstrations$^{\heartsuit}$             & 92.2 (1.4) & 51.2 (1.6) & 88.2 (0.9)                     & 91.8 (1.5) & 85.5 (4.2) & 90.9 (1.9) & \multicolumn{1}{l}{87.6 (4.8)} \\
\quad + demonstration-label re-prediction & 92.8 (0.7) & 51.4 (1.0) & 89.2 (1.0)                     & 92.2 (1.2) & 87.5 (1.0) & 92.1 (1.6) & \multicolumn{1}{l}{89.9 (3.1)} \\
\quad + contrastive learning          & 93.1 (0.5) & 52.3 (0.6) & 89.1 (1.0)                     & 91.8 (0.7) & 87.7 (1.2) & 92.4 (1.1) & 89.1 (3.2)                     \\ \hline
                               & \begin{tabular}[c]{@{}c@{}}MNLI\\ $($acc$)$ \end{tabular}       & \begin{tabular}[c]{@{}c@{}}MNLI-mm\\ $($acc$)$ \end{tabular}    & \begin{tabular}[c]{@{}c@{}}SNLI\\ $($acc$)$ \end{tabular}                           & \begin{tabular}[c]{@{}c@{}}QNLI\\ $($acc$)$ \end{tabular}       & \begin{tabular}[c]{@{}c@{}}RTE\\ $($acc$)$ \end{tabular}        & \begin{tabular}[c]{@{}c@{}}MRPC\\ $($F1$)$ \end{tabular}       & \begin{tabular}[c]{@{}c@{}}QQP\\ $($F1$)$ \end{tabular}                            \\ \hline
Majority                       & 32.7      & 33.0      & 33.8                          & 49.5      & 52.7      & 52.7      & 0.0                           \\
Prompt-based zero-shot         & 50.8      & 51.7      & 49.5                          & 50.8      & 51.3      & 61.9      & 49.7                          \\
“GPT-3” in-context learning    & 52.0 (0.7) & 53.4 (0.6) & 47.1 (0.6)                     & 53.8 (0.4) & 60.4 (1.4) & 45.7 (6.0) & 36.1 (5.2)                     \\ 
Fine-tuning    & 45.8 (6.4) & 47.8 (6.8) & 48.4 (4.8)                     & 60.2 (6.5) & 54.4 (3.9) & 76.6 (2.5) & 60.7 (4.3)                     \\
P-tuning    & 61.5 (2.1) & - & 72.3 (3.0) & 64.3(2.8)                     & - & 76.2 (2.3) & 65.6 (3.0)                      \\
DART  & 67.5 (2.6) & - & 75.8 (1.6)                     & 66.7 (3.7) & -  & 78.3 (4.5) & 67.8 (3.2)                     \\ 
Li's & 69.2 (4.0) & 71.0 (3.5)                     & 79.3 (3.2) & 69.0 (4.5)  & \textbf{74.2} (3.1) & 73.2 (7.5) & 68.2 (3.4) \\
EFL $^{\heartsuit}$ & 65.8 (3.7) & 68.5 (2.8)                     & 78.2 (1.3) & 67.6 (5.5)  & 68.9 (1.5) & 77.4 (6.3) & 67.0 (2.9)  \\
LM-BFF $^{\heartsuit}$ & 69.6 (2.9) & 71.3 (2.6) & \multicolumn{1}{l}{78.0 (3.6)} & 68.8 (5.4) & 68.7 (2.3)          & 77.3 (6.0) & 68.7 (4.7)                     \\
Imitation-Demo (ours) & \textbf{71.4} (0.9) & \textbf{72.0} (2.0) & \textbf{80.0} (3.3)                     & \textbf{70.5} (3.3) &  71.5 (1.5)         & \textbf{80.8} (3.2) & \textbf{70.9} (1.5)  \\
\hline 
Prompt-based Fine-tuning (man)  $^{\heartsuit}$          & 68.3 (2.3) & 70.5 (1.9) & \multicolumn{1}{l}{77.2 (3.7)} & 64.5 (4.3) & 69.1 (3.6)          & 74.5 (5.3) & 65.5 (5.3)                     \\
\quad + demonstrations$^{\heartsuit}$              & 69.6 (2.9) & 71.3 (2.6) & \multicolumn{1}{l}{78.0 (3.6)} & 68.8 (5.4) & 68.7 (2.3)          & 77.3 (6.0) & 68.7 (4.7)                     \\
\quad + demonstration-label re-prediction & 71.3 (0.9) & 72.5 (1.4) & \multicolumn{1}{l}{79.6 (3.2)} & 70.3 (4.1) & 70.8 (3.4)          & 77.0 (2.6) & 68.8 (2.6)                     \\
\quad + contrastive learning          & 71.4 (0.9) & 72.0 (2.0) & 80.0 (3.3)                     & 70.5 (3.3) &  71.5 (1.5)         & 80.8 (3.2) & 70.9 (1.5)                     \\ \bottomrule[1pt]
\end{tabular}
}
\caption{Overall results on RoBERTa-large with 16 samples per class. We report the mean (variance) of models trained on 5 different randomly sampled training and dev splits. Prompt-based Fine-tuning (man) indicates trained with manually designed templates. $^{\heartsuit}$ denotes we re-implement the EFL and LM-BFF models for fair comparisons. }
\label{main:exp}
\end{table*}

\section{Experiments}  
\textbf{Experiments Settings. } Experiments are conducted on 14 widely-used classification datasets. The datasets and evaluation metrics details are listed in Appendix \ref{appendix:Datasets}. Besides, the hyper-parameters settings of our experiments are listed in Appendix \ref{appendix:Parameters }. \\
\textbf{Compared Methods.} (1) \textbf{Majority}, which select the majority class of the dataset; (2) \textbf{Prompt-based zero-shot}: which use prompt tunning in zero-shot situations; (3) \textbf{``GPT-3'' in-context learning}, which use the in-context learning proposed in RoBERTa with no parameter updating; (4) \textbf{Fine-tuning}; (5) \textbf{P-tuning} \cite{DBLP:journals/corr/abs-2103-10385}, which employ trainable continuous prompt embeddings; (6) \textbf{DART} \cite{DBLP:journals/corr/abs-2108-13161}, which differentially optimize the prompt template and the target label during the backpropagation process; (7) \textbf{Li's} \cite{DBLP:journals/corr/abs-2203-03235}, which reformulate a classification or a regression task as a token-replaced detection problem utilizing pre-trained model Electra \cite{DBLP:conf/iclr/ClarkLLM20}; (8) \textbf{EFL} \cite{DBLP:journals/corr/abs-2104-14690}, which reformulate potential NLP task into an entailment one. (9) \textbf{LM-BFF} \cite{DBLP:conf/acl/GaoFC20}, which manually design templates and augment prompt tuning with demonstrations. \\
\textbf{Main Results. }
We can conclude from the experiment results illustrated in Table~\ref{main:exp} that: 
(1) The methods leveraging demonstrations (e.g.,  LM-BFF) generally achieve productive results, proving the superiority of demonstration learning mechanism.
(2) P-tuning and DART leverage continuous prompt embeddings to boost experiment results. Meanwhile, Li's and EFL promote performance by converting the cloze-style task into token-replaced detection or entailment problem.
However, without introducing any additional parameters or any prediction computation, Imitation-Demo achieves state-of-the-art results on 11 out of 14 datasets in the original mask-prediction way.
The performance gain indicates that Imitation-Demo could effectively promote experiment results by reinforcing the connections between prompt and demonstrations. 
(3) Ablation experiment results in the lower part of Table \ref{main:exp} illustrate the effectiveness of the contrastive learning and demonstrations-label re-prediction methods.
\begin{table}[]
\centering
\scalebox{0.8}{
\begin{tabular}{lcccc}
\toprule[1pt]
          & QQP & MNLI-mm &  MNLI  \\ \hline
LM-BFF    & 1.11 & 1.02    & 1.01 \\
Imitation-Demo & 1.16 & 1.04    & 1.05 \\ \bottomrule[1pt]
\end{tabular}
}
\caption{Averaged RoBERTa attention results pointing from demonstrations to prompt. The values are normalized by default RoBERTa pre-training weights.}
\label{test:attention}
\end{table}\\
\textbf{Analysis. }
Extensive experiments are conducted to show that our human-like imitation mechanisms enhance the connection between prompt and demonstration. Firstly, 
when trained with random demonstration labels, 
as shown in Table \ref{test:random}, we observe that Imitation-Demo has a greater drop rate than LM-BFF, indicating $[MASK]$ is dependent more on the semantic information from demonstrations. This finding could explain why there is little performance degradation when using random demonstration labels in \citet{DBLP:journals/corr/abs-2202-12837} to some extent. 
Moreover, following \citet{DBLP:conf/acl/WeiSZZGJ20}, we further conduct an experiment to show the review process with attention weights of the RoBERTa backbone. 
We average the total 384 attention heads of Roberta-large pointing from demonstrations to prompt, then normalize the values by default RoBERTa pre-trained weights. From the results in Table \ref{test:attention}, we observe Imitation-Demo received larger attention values. The result indicates that our approach could direct the RoBERTa model by modifying the corresponding attention weights and guiding prompt to focus more on the clues brought by demonstrations.

\section{Conclusion}
In this paper, we propose imitation demonstration learning to reinforce the correlations between prompt and given demonstrations. Inspired by the human review process, we introduce contrastive learning and demonstration-label re-prediction mechanisms to strengthen the two sub-steps of human review behaviour, respectively.
Experiments on 14 classification datasets show that our method consistently outperforms other baselines in few-shot settings. 
We hope this work could inspire the exploration of the working mechanism of demonstration learning and toward better few-shot learning abilities.

\bibliography{anthology,custom}
\bibliographystyle{acl_natbib}

\appendix

\section{Appendix}
\subsection{Datasets}
\label{appendix:Datasets}
Following the settings in \cite{DBLP:conf/acl/GaoFC20}, we evaluate on 14 classification datasets. For SNLI \cite{DBLP:conf/emnlp/BowmanAPM15}, SST-2 \cite{DBLP:conf/emnlp/SocherPWCMNP13}, CoLA \cite{DBLP:journals/tacl/WarstadtSB19}, MNLI \cite{DBLP:conf/naacl/WilliamsNB18}, QNLI \cite{DBLP:conf/emnlp/RajpurkarZLL16}, RTE \cite{DBLP:conf/mlcw/DaganGM05,giampiccolo2007third,bentivogli2009fifth}, MRPC \cite{DBLP:conf/acl-iwp/DolanB05}, QQP\footnote{https://www.quora.com/q/quoradata/} and SST-B \cite{DBLP:conf/semeval/CerDALS17}, we use the original development sets for testing. For MR  \cite{DBLP:conf/acl/PangL05}, CR \cite{DBLP:conf/kdd/HuL04}, MPQA \cite{DBLP:journals/lre/WiebeWC05} and Subj \cite{DBLP:conf/acl/PangL04}, we randomly sample 2,000 examples as the testing set. For SST-5 \cite{DBLP:conf/emnlp/SocherPWCMNP13} and TREC \cite{DBLP:conf/sigir/VoorheesT00}, we use the official test sets. F1 score (F1) are adopted as the evaluation metric of MRPC and QQP, and the other datasets utilize accuracy (acc) as the evaluation criteria.

\subsection{Parameters Setting}
\label{appendix:Parameters }
We implement all the baselines and our frameworks using PyTorch \cite{DBLP:conf/nips/PaszkeGMLBCKLGA19}. The pre-trained \textit{RoBERTa-large} model and \textit{roberta-large-nli-stsb-mean-tokens} SBERT \cite{DBLP:conf/emnlp/ReimersG19} from huggingface\footnote{https://github.com/huggingface/transformers} are applied in the experiments. We get 16 samples per class during training for all models. 
In order to control the smoothness of the exponential functions when calculation contrastive learning loss, we divide  every mean-pooling results with temperature $T$.
Grid search mechanisim are utilized to select optimal hyper-parameter combinations on each split (the coefficients of demonstration-label re-prediction $\alpha$ from \{0.5,1,5,10\}, contrastive learning $\beta$ from \{5,10\}, and the temperature $T$ of contrastive learning from \{5,10\}), Finally we select the 
the coefficients $\alpha$ and $\beta$ as 1 and 5, respectively. The temperature $T$ is set as 5 and the batch size is 16. The other hyper-parameters are identical to the default settings in LM-BFF \cite{DBLP:conf/acl/GaoFC20} for fair comparison. We report the average performance of models trained on 5 different randomly sampled training and dev splits, the random seeds are fixed as 13, 32, 42 ,87 , 100, respectively.\\

\end{document}